%% file: main.tex
\documentclass[10pt,twocolumn,letterpaper]{article}
\usepackage{caption}
\usepackage{graphicx}
\usepackage{cvpr}              
\usepackage{bbding}
\usepackage{tabularx}
\usepackage{amsmath}
\usepackage{amssymb}

\input{preamble}

\definecolor{cvprblue}{rgb}{0.21,0.49,0.74}
\usepackage[pagebackref,breaklinks,colorlinks,citecolor=cvprblue]{hyperref}

\def\paperID{00019} 
\def\confName{EarthVision}
\def\confYear{2024}

\title{GeoSynth: Contextually-Aware High-Resolution Satellite Image Synthesis}

\author{Srikumar Sastry, Subash Khanal, Aayush Dhakal, Nathan Jacobs\\
Washington Univerity in St. Louis\\
{\{\tt\small s.sastry, k.subash, a.dhakal, jacobsn\}@wustl.edu}
}

\begin{document}

\twocolumn[{%
\renewcommand\twocolumn[1][]{#1}%
\maketitle
\begin{center}
    \centering
    \captionsetup{type=figure}
    \includegraphics[width=\textwidth]{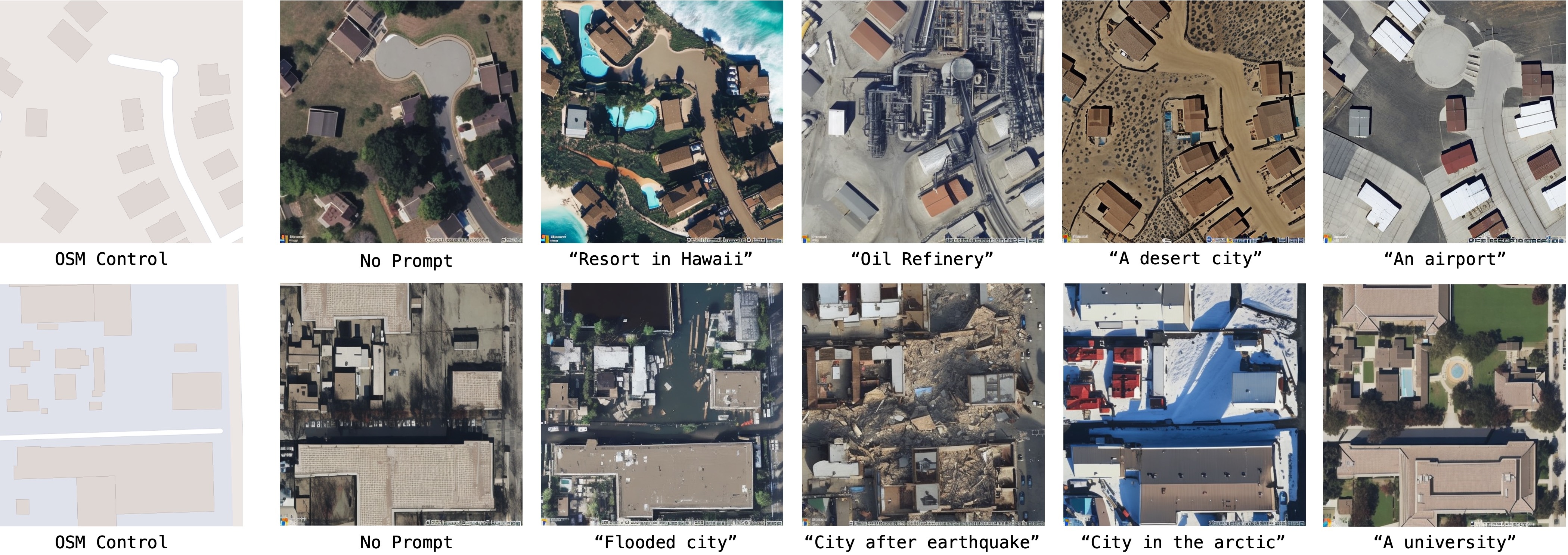}
    
    \captionof{figure}{Satellite images synthesized by GeoSynth using OpenStreetMap for layout control and textual prompts for style control. }
    \label{img:teaser}
\end{center}%
}]

\input{sec/0_abstract}    
\input{sec/1_intro}

\input{sec/2_related_works}
\input{sec/3_dataset}
\input{sec/4_method}
\input{sec/4_results}

{
    \small
    \bibliographystyle{ieeenat_fullname}
    \bibliography{main}
}
\input{sec/X_suppl}

\end{document}

%% file: preamble.tex
\usepackage[dvipsnames]{xcolor}


%% file: sec/0_abstract.tex
\begin{abstract}

We present GeoSynth, a model for synthesizing satellite images with global style and image-driven layout control. The global style control is via textual prompts or geographic location. These enable the specification of scene semantics or regional appearance respectively, and can be used together. 
We train our model on a large dataset of paired satellite imagery, with automatically generated captions, and OpenStreetMap data. We evaluate various combinations of control inputs, including different types of layout controls. Results demonstrate that our model can generate diverse, high-quality images and exhibits excellent zero-shot generalization. The code and model checkpoints are available at \url{https://github.com/mvrl/GeoSynth}. 






\end{abstract}

%% file: sec/1_intro.tex
\section{Introduction}
\label{sec:intro}

Imagine a scenario where you describe a scene and a layout and a realistic satellite image blooms into existence. Such kind of applications could assist in various remote sensing pipelines like urban planning, data augmentation, pseudo label generation for weakly supervised learning, etc. However, a single satellite image usually binds several spatial concepts into a unique image depicting complex and meaningful layouts. A scene on the ground, for example, may contain buildings, roads, intersections, crosswalks, trees, etc all placed together in a specific arrangement. This makes the problem of synthesizing realistic satellite images very challenging.

Recently, text-to-image models have rapidly redefined the realms of creativity and expression. When trained on vast datasets, they become capable of generating everything from photorealistic landscapes to fantastical creatures. In this regard, diffusion models~\cite{sohl2015deep,ho2020denoising,rombach2022high} have shown impressive performance in a variety of tasks such as image generation~\cite{saharia2022photorealistic,ramesh2022hierarchical}, image editing~\cite{hertz2022prompt,kawar2023imagic,brooks2023instructpix2pix}, video generation~\cite{gupta2023photorealistic,videoworldsimulators2024}, etc. 

Similarly, the field of remote sensing has witnessed remarkable progress in various aspects, including imaging technology, accessibility of high-resolution data, and global-scale applications. Thanks to the development of large-scale foundational models~\cite{cong2023satmae,reed2023scalemae,jakubik2023foundation,tseng2024lightweight,zhang2024earthgpt}, many remote sensing challenges have been tackled recently. The desirable properties of such foundational models have enabled domain-specific solutions in fields such as language~\cite{zavras2024mind,silva2024large}, sound~\cite{khanal2023learning}, natural images~\cite{dhakal2023sat2cap, cepeda2023geoclip}, etc. However, the majority of these machine learning-based methods 
fall short of utilizing the full potential of the satellite image modality~\cite{Rolf2024MissionC}. 
Along the same direction, less attention has been given to satellite image synthesis in remote sensing. Existing approaches to this problem are either application-specific~\cite{he2021spatial} or lack personalization capabilities~\cite{khanna2023diffusionsat}.

A fundamental difficulty in using the already existing diffusion models for synthesis is that diverse and high-resolution satellite images are unseen during their large-scale training. Furthermore, despite their general abilities, they fail to synthesize \textit{}{multiple specific} concepts in an image~\cite{kumari2023multi}. Recent lines of work addressing this challenge use a variety of techniques such as fine-tuning textual prompts~\cite{gal2022image}, end-to-end training~\cite{ruiz2023dreambooth}, manipulating the generation process~\cite{bar2023multidiffusion}, etc. ControlNet~\cite{zhang2023adding} has emerged as a promising architecture that learns to utilize information from an existing large-scale diffusion model while allowing it to condition it with a variety of controls. In this study, we use ControlNet to fine-tune Stable Diffusion (SD)~\cite{rombach2022high} to synthesize satellite images.

As shown in Figure~\ref{img:teaser}, this work aims to be able to synthesize realistic-looking satellite images, whose layout could be controlled via a reference image (for example OpenStreetMap (OSM) images), while style could be controlled using textual prompts. We additionally condition our models on geographic location using features extracted from SatCLIP~\cite{klemmer2023satclip}, a model trained contrastively on satellite images and geographic location. By doing so, our model exhibits synthesis capability conditioned on the geography of a region. In addition to the OSM control, we test two other conditioning controls: Canny edge and Segment Anything (SAM)~\cite{kirillov2023segment} mask, which can be directly obtained from raw satellite images. In the end, we have a suite of models namely, GeoSynth, which is capable of synthesizing satellite images that are optionally conditioned on layout, textual prompt and/or geographic location.
\noindent
The contributions of this work are threefold:
\begin{enumerate}
    \item We use features extracted from ControlNet and SatCLIP for high-resolution satellite image synthesis.
    \item We test the performance of three conditioning controls for synthesis: OSM image, Canny edge image, and Segment Anything mask.
    \item We demonstrate excellent zero-shot capabilities of our models.
\end{enumerate}

%% file: sec/2_related_works.tex
\section{Related Works}
\label{sec:related_works}

\subsection{Diffusion Models}
Sohl-Dickstein \etal \cite{sohl2015deep} first proposed physics-inspired generative models called diffusion probabilistic models to learn data distribution through parameterized Markov chains. Inspired by this, Ho \etal \cite{ho2020denoising} demonstrated that diffusion models can generate high-quality images. In the following years, Stable Diffusion \cite{rombach2022high} was introduced which proposed training in the latent space of pre-trained autoencoders leading to low computation cost and high-quality conditional image synthesis. Another competitive model, Imagen \cite{saharia2022photorealistic} utilizes a powerful large language model (LLM) to yield photorealistic images conditioned on text while training the diffusion model in pixel space. Motivated by the impressive results of these works, there has been an explosion of numerous diffusion models for controlled image synthesis \cite{ramesh2022hierarchical,epstein2024diffusion} leading to diverse capabilities such as image editing \cite{hertz2022prompt,kawar2023imagic,brooks2023instructpix2pix,xu2024personalized}, image stylization \cite{huang2024diffstyler,tumanyan2023plug,guo2023animatediff,sohn2024styledrop}, and video generation \cite{esser2023structure,gupta2023photorealistic,videoworldsimulators2024}. 

\subsection{Customization of Diffusion Models}
Leveraging the power of existing diffusion models pre-trained on large-scale datasets, two lines of work have emerged. The first focuses on developing methods to personalize the pre-trained diffusion model to encapsulate custom concepts and domains \cite{ruiz2023dreambooth,ruiz2023hyperdreambooth,kumari2023multi,chen2024subject,li2024blip,lu2023specialist,nitzan2023domain}. For instance, Dreambooth \cite{ruiz2023dreambooth} proposes personalization by learning from a small set of subject-specific images while preserving the prior of the pre-trained diffusion model. Other notable works propose techniques such as hypernetwork learning \cite{ruiz2023hyperdreambooth}, modifying the cross-attention layers \cite{kumari2023multi}, apprenticeship learning from a large number of concept-specific experts \cite{chen2024subject}, and utilizing the knowledge of existing multimodal representation space \cite{lu2023specialist,li2024blip}.
The second line of work focuses on developing training-free or efficient methods to incorporate different types of conditions into powerful diffusion models \cite{zhang2023adding,zhao2024uni,chen2024training,xie2023boxdiff,xue2023freestyle}. ControlNet~\cite{zhang2023adding} proposes zero-convolution-based modules to incorporate additional conditions such as Canny edge~\cite{canny1986computational}, sketch, pose, etc. Uni-ControlNet \cite{zhao2024uni} proposes to train separate adapters for two sets of controls: local controls (e.g., segmentation masks) and global controls (e.g., CLIP embeddings). Apart from these, a few recent works have focused on spatial layout-controlled image synthesis~\cite{chen2024training,xue2023freestyle,xie2023boxdiff}. Inspired by these works, we utilize the spatial layout present in OSM images as one of our conditions while employing the flexible approach of ControlNet for our conditional image synthesis.
\subsection{Satellite Image Synthesis}
The literature on conditional satellite image synthesis is limited. Most of the previous works are focused on task-specific satellite image synthesis. For example, a recent work, EDiffSR \cite{xiao2023ediffsr} proposes to utilize diffusion models for the task of super-resolution of remote sensing images. Chen \etal \cite{chen2023spectraldiff} demonstrate the utility of features from diffusion model trained on hyperspectral remote sensing imagery, for the task of pixel-wise semantic segmentation. For the task of text-to-satellite image synthesis, \cite{xu2023txt2img} proposes a two-stage framework. First, a VQVAE-like framework \cite{van2017neural} is trained to learn a codebook of visual representations for satellite imagery. Second, text-conditioned prototypes are learned to be utilized by VQVAE decoder to synthesize an image from text.
A parallel work, DiffusionSat \cite{khanna2023diffusionsat} learns satellite image generation conditioned on freely available metadata and sparsely available textual description. They demonstrate the impressive performance of their model on various downstream tasks such as super-resolution and in-painting. Different from their work, we use detailed textual descriptions along with spatial layouts for satellite image synthesis. We propose to incorporate all the conditioning modalities through ControlNet, hence preserving the existing knowledge base of stable diffusion. Additionally, we use a foundational location encoder model: SatCLIP, to incorporate geographic location for the synthesis process.

%% file: sec/3_dataset.tex
\section{Background}
\textbf{Diffusion Models}. Diffusion Models~\cite{sohl2015deep,ho2020denoising} are a class of probabilistic models that learn to sample from a data distribution ($\mathcal{D}$) given numerous samples from that distribution. This is done by learning to denoise a variable sampled from a known prior noise distribution in a markov chain process. A popular choice for this noise distribution is the standard normal distribution. During training, given a noisy image $x_t$ at timestep $t$, the objective of the diffusion model ($\epsilon_\theta$) is to predict the noise added at timestep $t-1$ to obtain that image. The criterion is given by:
\begin{equation}
    \mathbb{E}_{x,c,t,\epsilon}\left[||\epsilon - \epsilon_\theta(x_t,c,t)||^2_2\right]
\end{equation}
where $x\sim\mathcal{D}$, $c$ is a conditioning modality, $t\in\{T,T-1,\ldots,\delta\}$ and $\epsilon \sim \mathcal{N}(0,I)$. $c$ can be text, raw image, segmentation map, etc.

\noindent
\textbf{Latent Diffusion Models (LDMs)}. Since the dimension of the original data distribution in the case of images is high, the diffusion process is computationally expensive. LDMs~\cite{rombach2022high} proposed to first encode the original samples into a low dimensional latent space and then perform the diffusion on the latent representations of the original samples. After a series of denoising steps, a decoder is used to reconstruct the original image from its latent representation.

\noindent
\textbf{ControlNet}. ControlNet~\cite{zhang2023adding} is used to add additional conditioning controls to an existing neural network without having to fine-tune the original network. This is done by transforming the feature maps extracted from the existing neural network, into a feature that is conditioned on a given control. Each block of the ControlNet is connected to a zero-initialized layer, which ensures no noise is added during training. 


\begin{figure}[t]
  \centering
   \includegraphics[width=\linewidth]{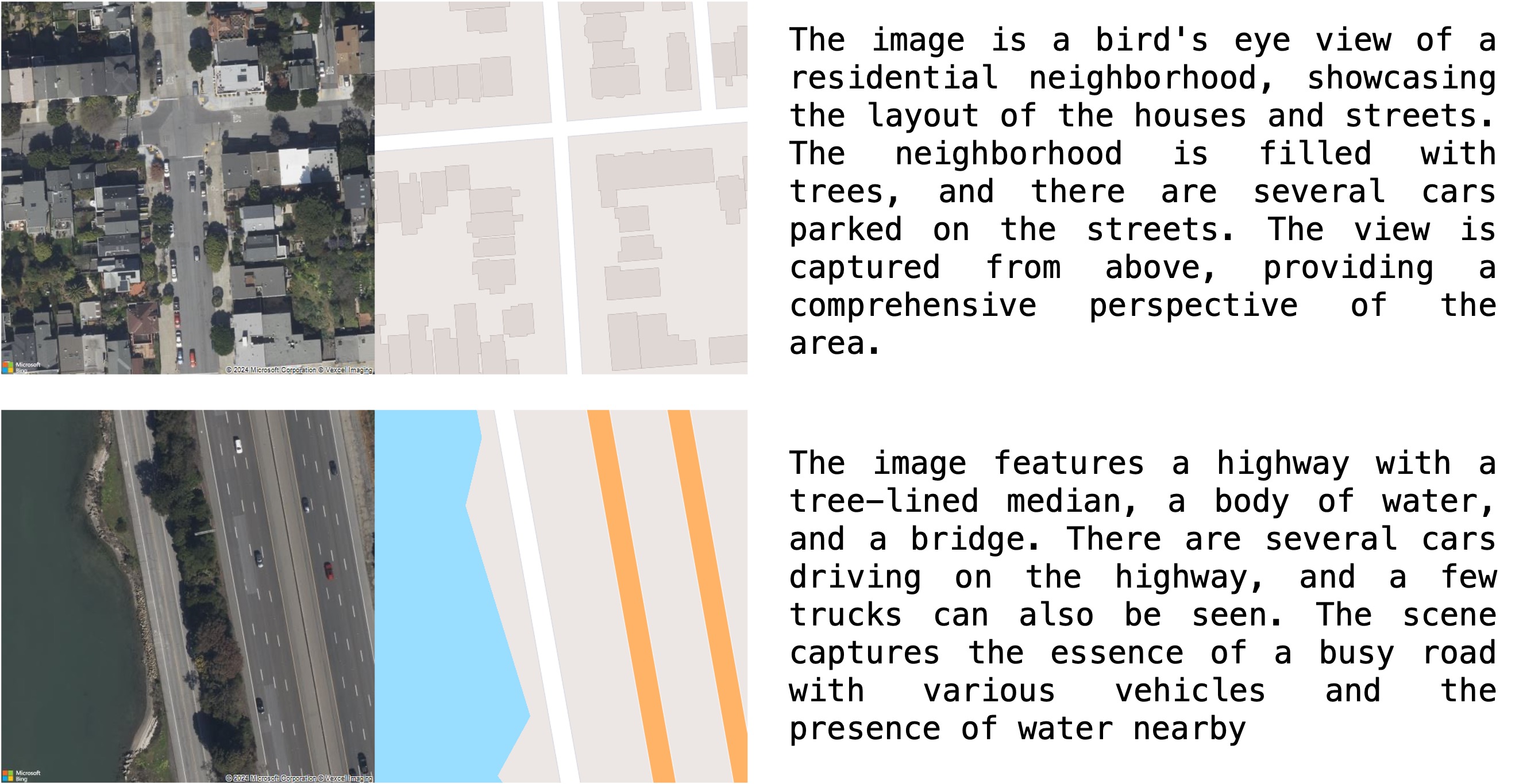}
   \caption{Each dataset sample consists of a satellite image, an OSM image, and an automatically generated textual description. Additionally, the dataset includes SAM masks for each satellite image.}
   \label{fig:dataex}
\end{figure}

\section{Dataset}

We built a dataset consisting of paired high-resolution satellite imagery and OSM images. We use a static representation of OSM in the form of images. The pairs were sampled randomly near ten major US cities, as shown in Appendix~\ref{dataset_sample}. To improve coverage and reduce spatial bias, each sampling location is spaced at least 1 kilometer apart from one another. All the images downloaded are of size 512x512 pixels at an approximate ground sampling distance of 0.6m. 
We downloaded 90,305 image pairs and filtered pairs consisting entirely of bare Earth, water, or forest. After filtering, the dataset contained 44,848 pairs. 

We extended the dataset by captioning each satellite image using LLaVA~\cite{liu2024visual}, a recently released multimodal large language model (Figure~\ref{fig:dataex}). The prompt used for captioning was: ``Describe the contents of the image". The captioning pipeline took 40 GPU hours to run on 2 NVIDIA A6000 GPUs. Lastly, we extracted the Canny edge image and the Segment Anything mask corresponding to each satellite image.

%% file: sec/4_method.tex
\begin{figure*}[t]
  \centering
   \includegraphics[width=\linewidth]{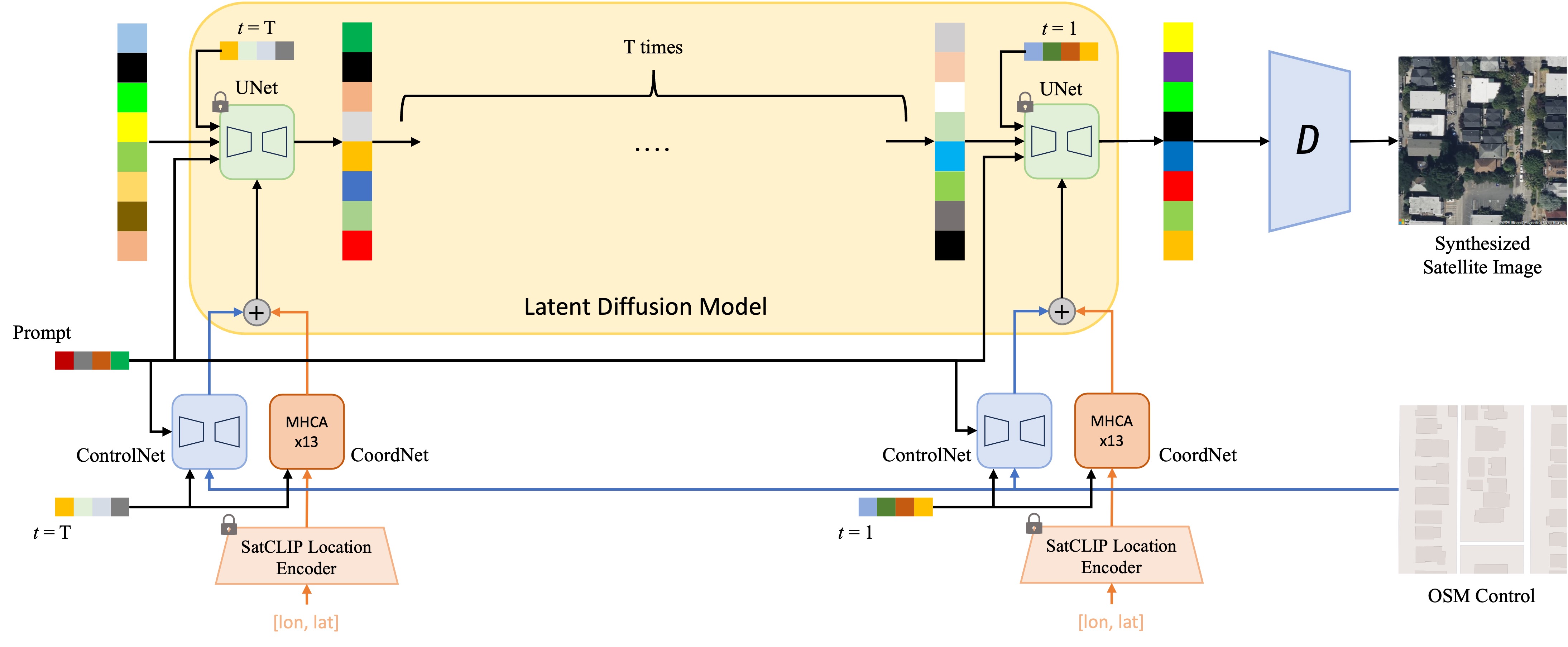}
   \caption{A high-level architecture overview of GeoSynth, which consists of a pre-trained LDM, ControlNet and CoordNet.}
   \label{fig:arc}
\end{figure*}


\section{Method}
Our goal is to train a suite of models that are capable of synthesizing satellite images ($x$) given a text prompt ($\tau$), geographic location ($l$), and a control image ($c$). This is done by training diffusion models to learn the conditional distribution $p(x|\tau,l,c)$. To this end, we use Latent Diffusion Models (LDM), which have shown state-of-the-art performance in conditional image synthesis. Below, we describe the model architecture used and the implementation details of the training pipeline.


\subsection{Architecture}

We utilize a pre-trained LDM that comprises four primary architectural components. Firstly, an encoder that transforms raw images into a low-dimensional latent space. Secondly, a pre-trained CLIP text encoder~\cite{radford2021learning} processes the raw text prompts and generates latent text vectors. Thirdly, the diffusion model has a U-Net based architecture consisting of cross-attention blocks. Finally, a decoder reconstructs images given their corresponding latent vectors. The encoder and decoder of th LDM have a Variational Autoencoder (VAE) style architecture. During training, the diffusion process is used in the latent representation space of the raw images. The diffusion model learns to denoise a noisy latent vector at a given timestep conditioned on the text prompt. 

As shown in Figure~\ref{fig:arc}, we use ControlNet to incorporate a control image and fine-tune the pre-trained LDM. ControlNet is a zero-initialized neural network attached on top of an LDM, which transforms the feature maps of the LDM at each stage. 
The ControlNet architecture consists of 13 residual cross-attention blocks which take as input the control image, text prompt, and the diffusion timestep. 

To incorporate geographic location as a condition, we first use SatCLIP~\cite{klemmer2023satclip} to extract location-based features. SatCLIP is a spherical harmonics-based location encoder that provides general-purpose location embeddings. It is trained using a contrastive learning framework with a CLIP-style satellite image encoder. We design a ControlNet-style cross-attention-based transformer, namely CoordNet, which processes the location embeddings. CoordNet consists of 13 layer multi-head cross-attention blocks which take as input the SatCLIP location-based embeddings and the diffusion timestep. Each cross-attention block in the CoordNet consists of a zero-initialized feed-forward layer. The features extracted from each of its blocks are added to the features extracted from the corresponding blocks of the ControlNet. These features are then added to the corresponding residual blocks of the LDM. During training, all the LDM components and the SatCLIP location encoder are frozen, as shown in Figure~\ref{fig:arc}.

During inference, a noisy latent vector is sampled from a standard normal distribution. The diffusion model is then used to progressively denoise the latent vector over a series of T timesteps. The inputs from the CLIP text encoder, the ControlNet, and the CoordNet are used to guide the denoising process at each timestep.
\begin{figure*}
  \centering
  \includegraphics[width=\linewidth]{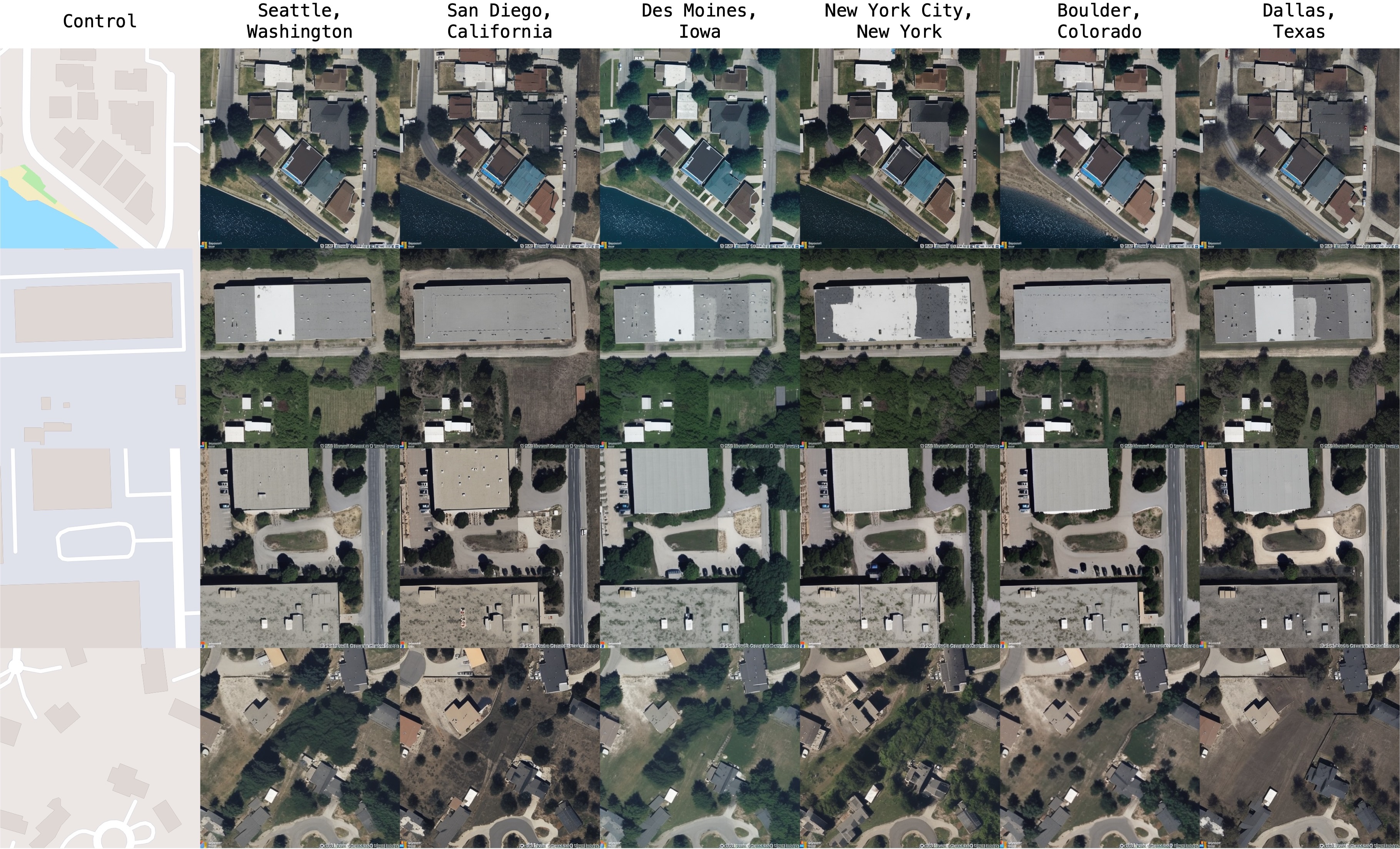}
  \caption{Geo-aware generation. We show four example generations of satellite images using six different geographic locations. We use the same OSM control and random seed without specifying any textual prompt.}
  \label{fig:loc_sample}
\end{figure*}
\begin{table*}
  \centering
  \begin{tabularx}{0.74\linewidth}{lcccccc}
    \toprule
    Method & Control & Location & FID $\downarrow$ & SSIM $\uparrow$ & CLIP-Score $\uparrow$ \\
    \midrule
    GeoSynth & - & - &13.55 &0.237 &0.287 \\
    GeoSynth & - & \Checkmark & 12.01 (\textbf{+1.54}) & 0.264 (\textbf{+0.027}) & 0.288 (\textbf{+0.001})\\
    \bottomrule
    \bottomrule
    GeoSynth & Canny Edge & - & 15.35 & 0.350&0.291\\
    GeoSynth& Canny Edge & \Checkmark & 13.92 (\textbf{+1.43}) &0.361 (\textbf{+0.011}) &0.289 (-0.002) \\
    \bottomrule
    \bottomrule
    GeoSynth& SAM Mask & - & 12.29 & 0.335& 0.297 \\
    GeoSynth& SAM Mask & \Checkmark & 12.04 (\textbf{+0.25}) & 0.346 (\textbf{+0.011}) & 0.290 (-0.007) \\
    \bottomrule
    \bottomrule
    GeoSynth& OSM & - &12.97 & 0.274 & 0.298\\
    GeoSynth& OSM & \Checkmark &11.90 (\textbf{+1.07}) &0.291 (\textbf{+0.017})  &0.303 (\textbf{+0.005}) \\
    \bottomrule
  \end{tabularx}
  \caption{Incorporating geographic location as an additional condition results in higher FID and SSIM scores. For each of these experiments, we incorporated text prompts during the training. }
  \label{tab:location}
\end{table*}
\subsection{Implementation Details}
We use Stable Diffusion (SD) v2.1 as the pre-trained LDM. We use the same base architecture of ControlNet as used by authors of~\cite{zhang2023adding}. CoordNet consists of stacked 13 cross-attention blocks with an inner dimension of 256 and 4 heads. To improve training speeds, we precomputed the location-based embeddings of SatCLIP for each of the images and saved them on disk. 

In total, we trained 14 variants of the model, including and excluding the conditioning modalities. Each variant of the model is trained on 2 NVIDIA RTX 4090 for a total of 100 GPU hours. We train the models using the DistributedDataParellel routine in PyTorch. We use the Adam optimizer with a learning rate of $1e^{-5}$, gradient accumulation over 16 batches, and a batch size of 4 on each GPU.
Following~\cite{zhang2023adding}, we randomly mask out the textual prompts with a probability of 0.5 during training. This ensures that the model learns the semantic information present in the control images independent of the textual prompts.

We use three distinct metrics to evaluate the performance of the models. The first metric is the Frechet Inception Distance (FID)~\cite{heusel2017gans}, which indicates the distance between the synthesized and ground-truth data distribution at the feature level. The second metric is SSIM~\cite{wang2004image}, which measures the similarity between synthesized images and ground-truth samples at the pixel level. Lastly, we use the CLIP-score~\cite{radford2021learning} to measure the similarity between synthesized images and corresponding text prompts.


%% file: sec/4_results.tex
\section{Results and Discussion}
\label{sec:formatting}
\begin{figure*}
  \centering
  \includegraphics[width=\linewidth]{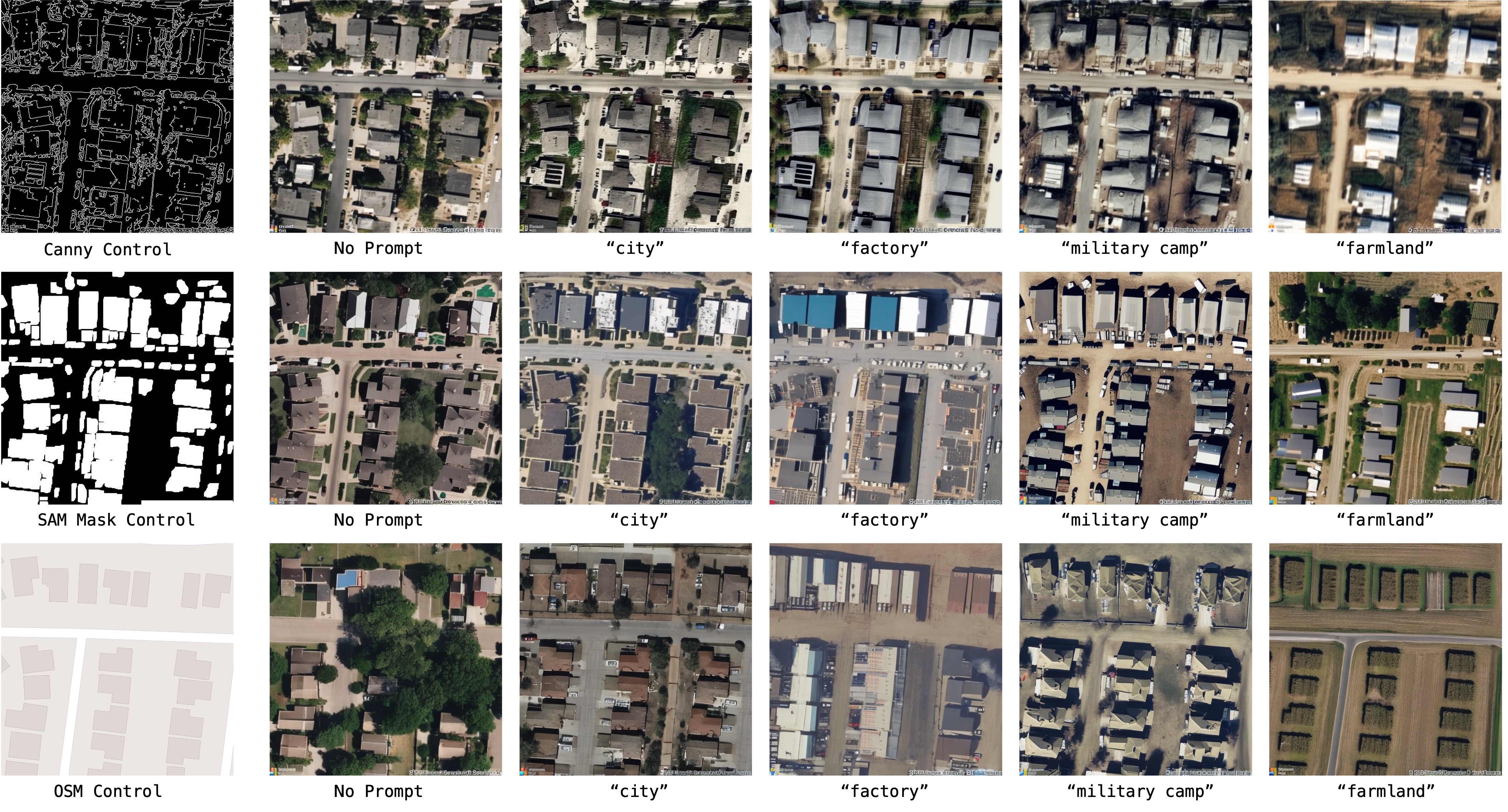}
  \caption{Synthesis performance of GeoSynth when using various layout controls and text prompts.}
  \label{fig:control_sample}
\end{figure*}

\begin{table*}
  \centering
  \begin{tabularx}{0.70\linewidth}{lcccccc}
    \toprule
    Method & Control & Text & FID $\downarrow$ & SSIM $\uparrow$ & CLIP-Score $\uparrow$ \\
    \midrule
    GeoSynth & - & - &16.11 & 0.199 &0.207 \\
    GeoSynth & - & \Checkmark &13.55 (\textbf{+2.56}) &  0.237 (\textbf{+0.038}) &0.287 (\textbf{+0.080}) \\
    \bottomrule
    \bottomrule
    GeoSynth & Canny Edge & - & 16.74 & 0.200 & 0.274\\
    GeoSynth& Canny Edge & \Checkmark & 15.35 (\textbf{+1.39}) & 0.350 (\textbf{+0.15}) &0.291 (\textbf{+0.017}) \\
    \bottomrule
    \bottomrule
    GeoSynth& SAM Mask & - & 13.48 & 0.268 & 0.262 \\
    GeoSynth& SAM Mask & \Checkmark & 12.29 (\textbf{+1.19}) & 0.335 (\textbf{+0.067})& 0.297 (\textbf{+0.035})\\
    \bottomrule
    \bottomrule
    GeoSynth& OSM & - &12.70 & 0.273 & 0.269\\
    GeoSynth& OSM & \Checkmark & 12.97 (-0.27) & 0.274 (\textbf{+0.001}) & 0.298 (\textbf{+0.029})\\
    \bottomrule
  \end{tabularx}
  \caption{Text-guided training improves the quality and diversity of synthesis.}
  \label{tab:text}
\end{table*}

\begin{figure}
  \centering
  \includegraphics[width=\linewidth]{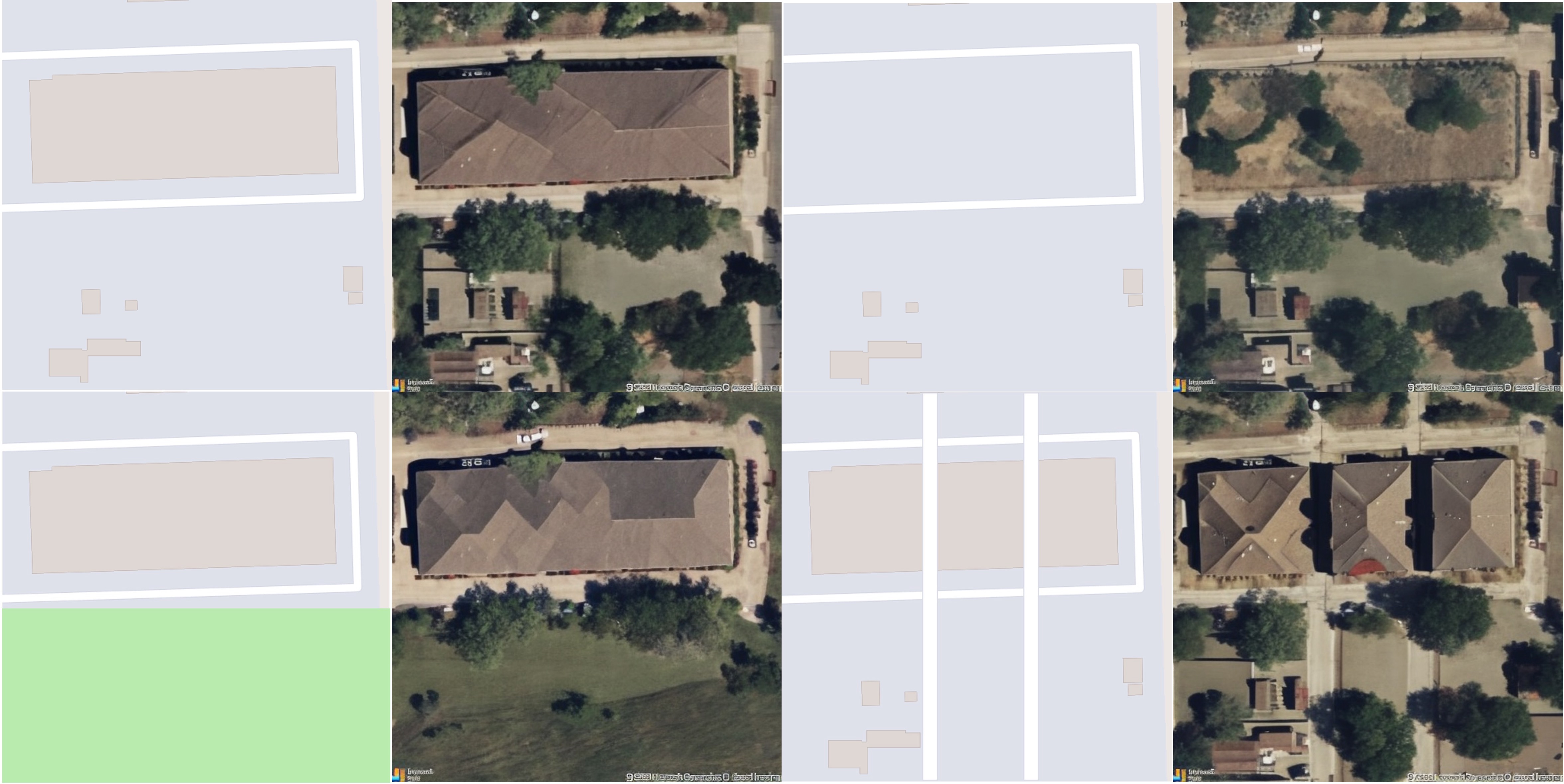}
  \caption{Editing. We show three example generations of satellite images using edited versions of a control OSM image. We use the same random seed without specifying any textual prompt. }
  \label{fig:edit}
\end{figure}

\textbf{Geo-aware synthesis}. In Figure~\ref{fig:loc_sample}, we demonstrate four example syntheses from varying geographic locations. We use the same control OSM image and random seed for each example. Additionally, no prompt was provided for synthesizing the images. It is observed that our models have learned high-level semantics of various geographic locations across the USA. This is confirmed by qualitatively examining the synthesized images, where Iowa is characterized by more greenery while California has more desert-like features. Geographic locations in and around New York tend to produce satellite images with heavy urban development. Table~\ref{tab:location} presents a quantitative evaluation of the models when incorporating geographic location as an additional condition. Adding geographic location improves the ability of the models to create satellite images that look more realistic, as confirmed by better FID and SSIM scores. However, the CLIP-score shows little to no improvement, which is expected since this score measures the similarity between image and the corresponding text prompt. Text-only GeoSynth achieves a high FID and CLIP-score while receiving a poor SSIM score. This indicates that the model can produce semantically meaningful images that differ from the original ground-truth distribution at the pixel level.\\

\noindent
\textbf{Control Images}. In Figure~\ref{fig:control_sample}, we show satellite images synthesized using three controls: Canny edge image, SAM mask, and OSM image. We show result synthesis when using various challenging prompts. It is noticed that the model using Canny edge as a control generates the most realistic-looking satellite images, as confirmed by the high SSIM score. However, it is incapable of regulating the style of the satellite image as given in the prompt. Although SAM mask achieves the highest scores on average over all the metrics, it fails to produce visually aesthetic-looking satellite images. This happens due to the over-segmented masks produced by SAM when applied to satellite images. As proven by the highest FID and CLIP-score, OSM imagery as control produces the most semantically meaningful satellite images. Furthermore, the model shows a good zero-shot synthesis capability. It can effectively control the style of satellite images according to the prompt. As depicted in Figure~\ref{fig:edit}, our model can be utilized to edit satellite images by providing edited copies of OSM imagery. This is possible by using the same textual prompt and random seed.\\

\noindent
\textbf{Importance of text}. We experimented to determine the significance of text guidance in satellite image synthesis, as shown in Table~\ref{tab:text}. The models were trained both with and without text prompts, disregarding geographic location. In instances where we trained models without text guidance, we provided an empty string as the text prompt. Our findings indicate that the performance of GeoSynth is poor when it is trained without including any text. When text is not incorporated, the model is incapable of generating diverse images. However, by including text, GeoSynth can produce realistic-looking satellite images. Similar observations are made when models are additionally trained with control images. Across all the metrics, a significant gain is seen in the CLIP-score. This is expected since CLIP-score reflects the similarity between a textual prompt and the corresponding synthesized image.\\

\begin{table}
  \centering
  \begin{tabular}{lcc}
    \toprule
    Class & CLIP-Confidence $\uparrow$\\
    \midrule
    airport & 55.41 \\
    amusement park & 57.90 \\
    beach & 92.74 \\
    botanical garden & 56.94 \\
    factory & 78.78 \\
    farmland & 91.81 \\
    golf course & 87.35 \\
    harbor & 52.27 \\
    parking lot & 92.46 \\
    railway station & 70.94\\
    \bottomrule
  \end{tabular}
  \caption{CLIP zero-shot classification performance on the synthesized samples generated using GeoSynth with a fixed OSM image as control.}
  \label{tab:zero-shot}
\end{table}

\begin{figure}
  \centering
  \includegraphics[width=\linewidth]{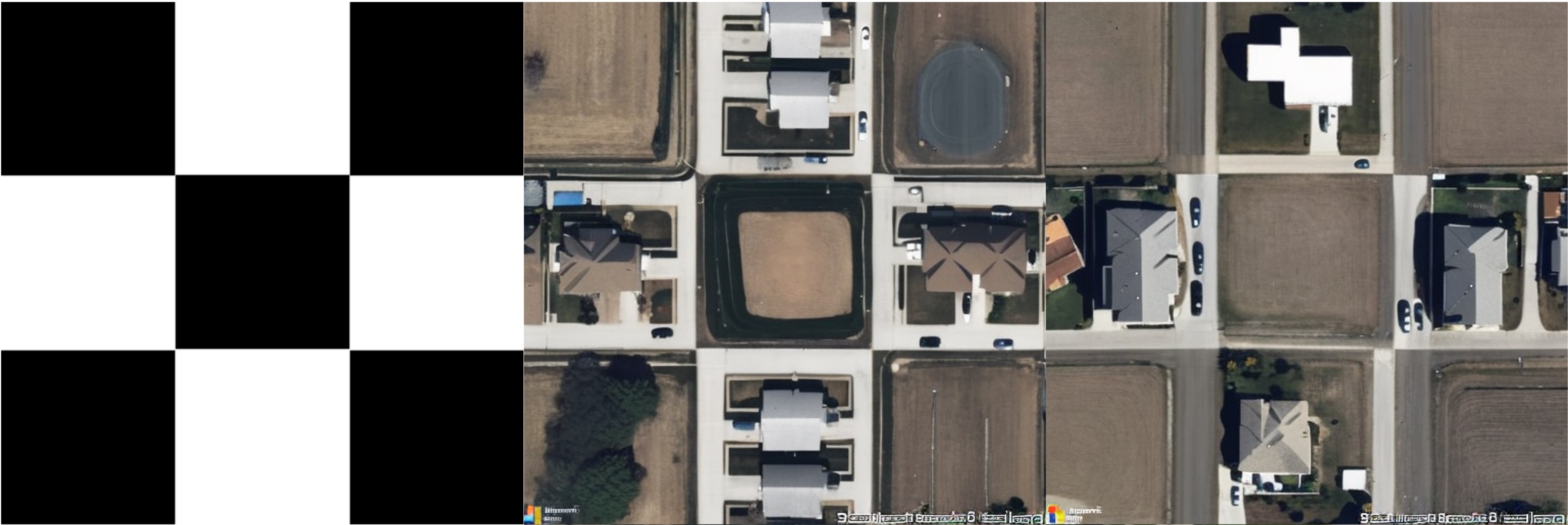}
  \caption{Example synthesis of GeoSynth using out-of-domain control image.}
  \label{fig:short}
\end{figure}

\begin{figure*}[!ht]
  \centering
  \includegraphics[width=\linewidth]{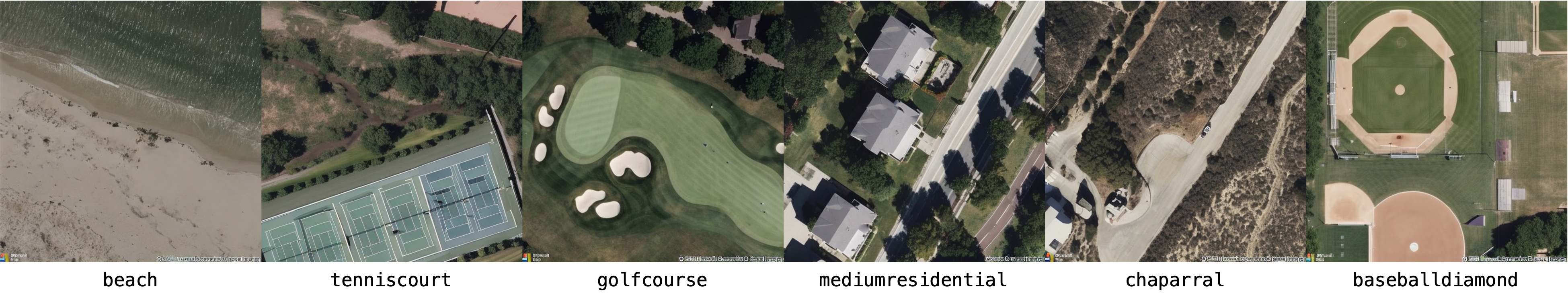}
  \caption{Generated samples from our GeoSynth model, without layout control, on UCMerced classes.}
  \label{fig:sd2.1}
\end{figure*}

\begin{table*}
  \centering
  \begin{tabular}{lcccccccc}
    \toprule
    Class & \multicolumn{2}{c}{GeoSynth (No Control)} & \multicolumn{3}{c}{GeoSynth (Canny Edge)} & \multicolumn{3}{c}{GeoSynth (SAM Mask)}\\
    & FID $\downarrow$ & CLIP-Score $\uparrow$ &FID $\downarrow$ & SSIM $\uparrow$ & CLIP-Score $\uparrow$ & FID $\downarrow$ & SSIM $\uparrow$ & CLIP-Score $\uparrow$ \\
    \midrule
    agricultural & 33.95 & 0.261 & 29.56 & 0.056 &0.245 &38.17&0.013&0.246\\
    airplane & 32.44 & 0.245 & 50.57 &0.090&0.227&70.19&0.087&0.222\\
    baseballdiamond & 22.57 & 0.307& 42.97 &0.089&0.276&48
    .81&0.185&0.280 \\
    beach & 32.94 &0.237 &39.07&0.271&0.214&43.73&0.200&0.216\\
    buildings & 26.97&0.245 &33.63&0.120&0.253&56.23&0.084&0.247 \\
    chaparral & 55.46&0.222 &46.04&0.123&0.216&51.72&0.088&0.206 \\
    denseresidential & 26.80 &0.267 &26.15&0.113&0.261&51.39&0.058&0.252\\
    forest & 18.48&0.265 &32.35&0.091&0.245&37.24&0.045&0.241 \\
    freeway & 22.86& 0.258&36.61&0.086&0.265&65.60&0.071&0.243 \\
    golfcourse &  23.88&0.274 &39.66&0.209&0.261&44.18&0.177&0.262\\
    harbor & 69.26&0.203 &29.32&0.137&0.224&38.90&0.074&0.230\\
    intersection & 27.25&0.271 &22.78&0.152&0.272&37.73&0.101&0.275\\
    mediumresidential & 20.13&0.254 &35.98&0.109&0.252&69.79&0.055&0.238\\
    mobilehomepark & 35.22& 0.287&50.79&0.109&0.257&61.74&0.048&0.243\\
    overpass & 21.61&0.262 &47.40&0.078&0.263&65.61&0.080&0.247\\
    parkinglot & 36.37&0.278 &34.70&0.097&0.256&49.66&0.060&0.243\\
    river & 21.63& 0.237&53.68&0.119&0.223&62.66&0.083&0.219\\
    runway & 45.78 &0.230&53.64&0.100&0.214&57.38&0.084&0.214\\
    sparseresidential & 36.49&0.262 &36.95&0.125&0.240&48.16&0.104&0.238\\
    storagetanks & 33.12&0.285 &55.71&0.125&0.233&64.89&0.103&0.243\\
    tenniscourt & 23.37&0.296 &57.01&0.101&0.238&50.90&0.097&0.257\\
    \bottomrule
  \end{tabular}
  \caption{The performance of zero-shot synthesis of our models on UCMerced categories.}
  \label{tab:ucmerced}
\end{table*}

\noindent
\textbf{Zero-shot capabilites}. We evaluated zero-shot generalization of our models. Firstly, we show the synthesis performance of our model when using out-of-domain control images. In Figure~\ref{fig:short}, our model was provided with a control image in the form SAM mask. Visually, the model has synthesized realistic-looking satellite images. Similar behavior is observed when the model is provided out-of-domain OSM or Canny edge image. Next, we assessed the performance of our model in generating a variety of concepts while using a fixed control image. We selected ten land-use categories and synthesized 50 images for each category, using fixed OSM imagery. We provided the names of these categories in the textual prompt and maintained a consistent random seed throughout the generation process. We employed CLIP's zero-shot classification pipeline to classify each synthesized image into a set of binary classes. For each category, we determined whether the generated images belonged to that category or not. In Table~\ref{tab:zero-shot}, we report the average confidence value of CLIP for each category. A higher score indicates that CLIP classified a synthesized image into a given land-use category with high confidence. Our results indicate that the generated images effectively represent the specified land-use categories. Across all the land-use categories, our model was able to achieve an average CLIP-confidence of 73.66, which indicated an image generated using our model depicted the correct land-use category 73.66\% times on average. Lastly, we evaluated the model's performance in synthesizing images of categories specified in the UCMerced dataset~\cite{yang2010bag}. UCMerced contains satellite images at 0.3m resolution across 21 land-use categories. Figure~\ref{fig:sd2.1} depicts the zero-shot synthesis capability of text-only GeoSynth model.  Table~\ref{tab:ucmerced} demonstrates the class-wise performance of our models on the UCMerced dataset. Overall, the models perform well in certain categories such as beaches, buildings, etc. On the other hand, they perform poorly in categories such as storage tanks, airplanes, etc. We observe that the performance of GeoSynth with Canny Edge image or SAM mask as layout control on UCMerced depends highly on the quality of the layout image itself. Most often, SAM produces undersegmented images when applied on UCMerced. On the other hand, the Canny algorithm produces a lot of false edges. In the future, it is possible to improve the overall performance by finetuning the models on additional datasets.\\



\section{Conclusions}
Text-to-image models have exhibited impressive performance and have been widely used in various end-user applications. However, there has been little to no research conducted on this topic in the field of remote sensing. This lack of adoption of such models in remote sensing represents a missed opportunity for building innovative applications. Therefore, we aim to inspire the remote sensing community and promote future research directions in conditional satellite image synthesis through this work. While we leave potential applications of our framework as future work, we believe that urban planners will benefit the most from it. One could imagine using our framework for automatic digital twin generation, urban growth simulation, and city planning. To encapsulate, we proposed GeoSynth, a suite of models capable of synthesizing realistic-looking satellite images while allowing personalization through text prompts. It uses spatial layout from input control images to guide the synthesis process. Additionally, our model incorporates geographic location as a condition that improves the synthesis quality by considering a region's geographical features. We hope GeoSynth represents the first step towards a global geography-aware synthesis model.

\section{Acknowledgements}
We thank Dr. Joseph O'Sullivan, Wiete Fehner, and others for useful discussions as part of WashU ESE 5934. The code for this project is highly inspired from Zhang \etal ~\cite{zhang2023adding} GitHub repository: \url{https://github.com/lllyasviel/ControlNet}.

%% file: sec/X_suppl.tex
\clearpage
\setcounter{page}{1}
\twocolumn[{%
\renewcommand\twocolumn[1][]{#1}%
\maketitlesupplementary
\appendix
\flushleft
\section{Qualitative Results on UCMerced}
\begin{center}
    \centering
    \captionsetup{type=figure}
    \includegraphics[width=0.72\linewidth]{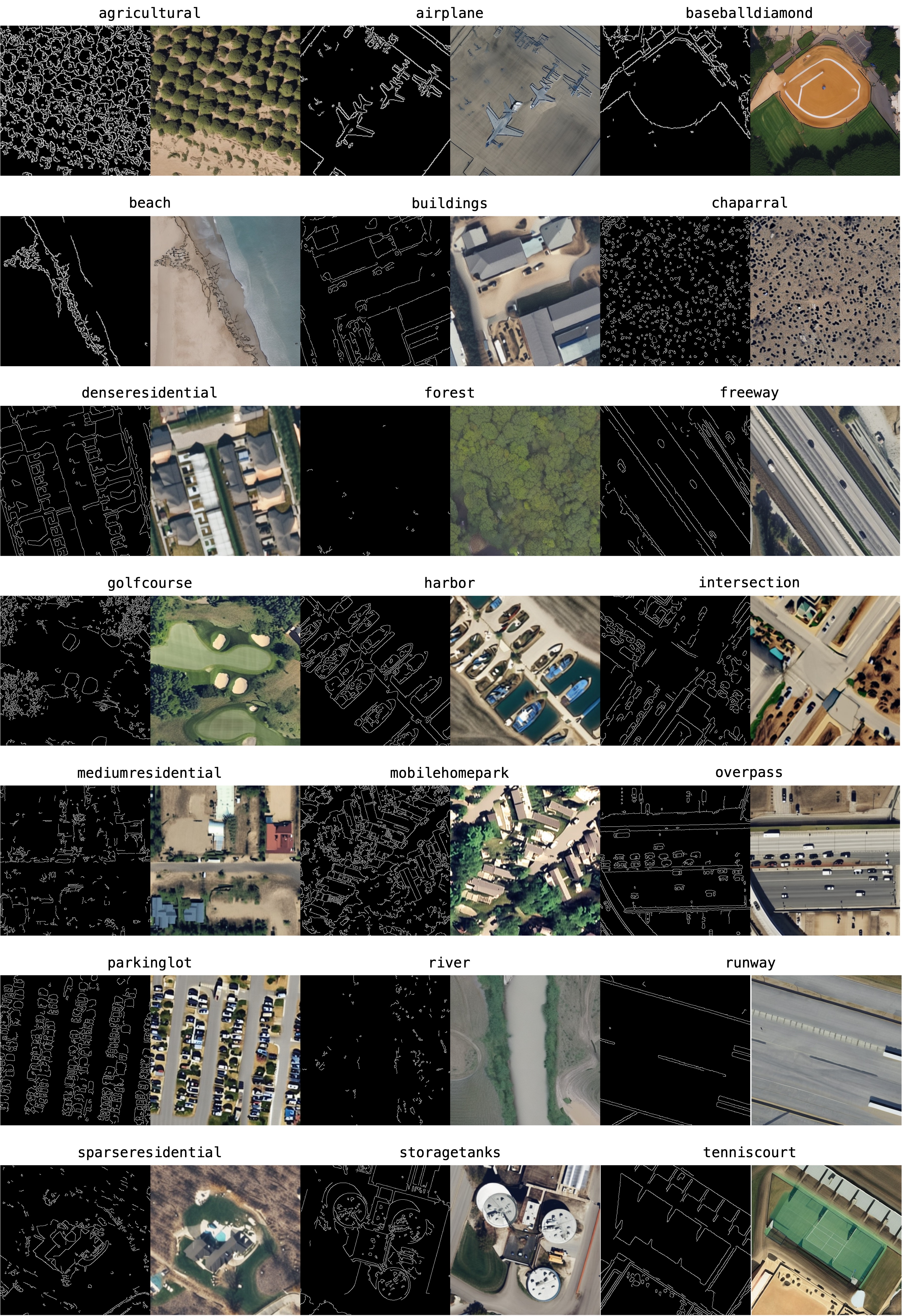}
    
    \captionof{figure}{Zero-shot synthesis using Canny edges. Examples of synthesized satellite images using Canny edges generated on UCMerced dataset. }
    \label{fig:supl_1}
\end{center}%
}]
\twocolumn[{%
\renewcommand\twocolumn[1][]{#1}%
\begin{center}
    \centering
    \captionsetup{type=figure}
    \includegraphics[width=0.82\linewidth]{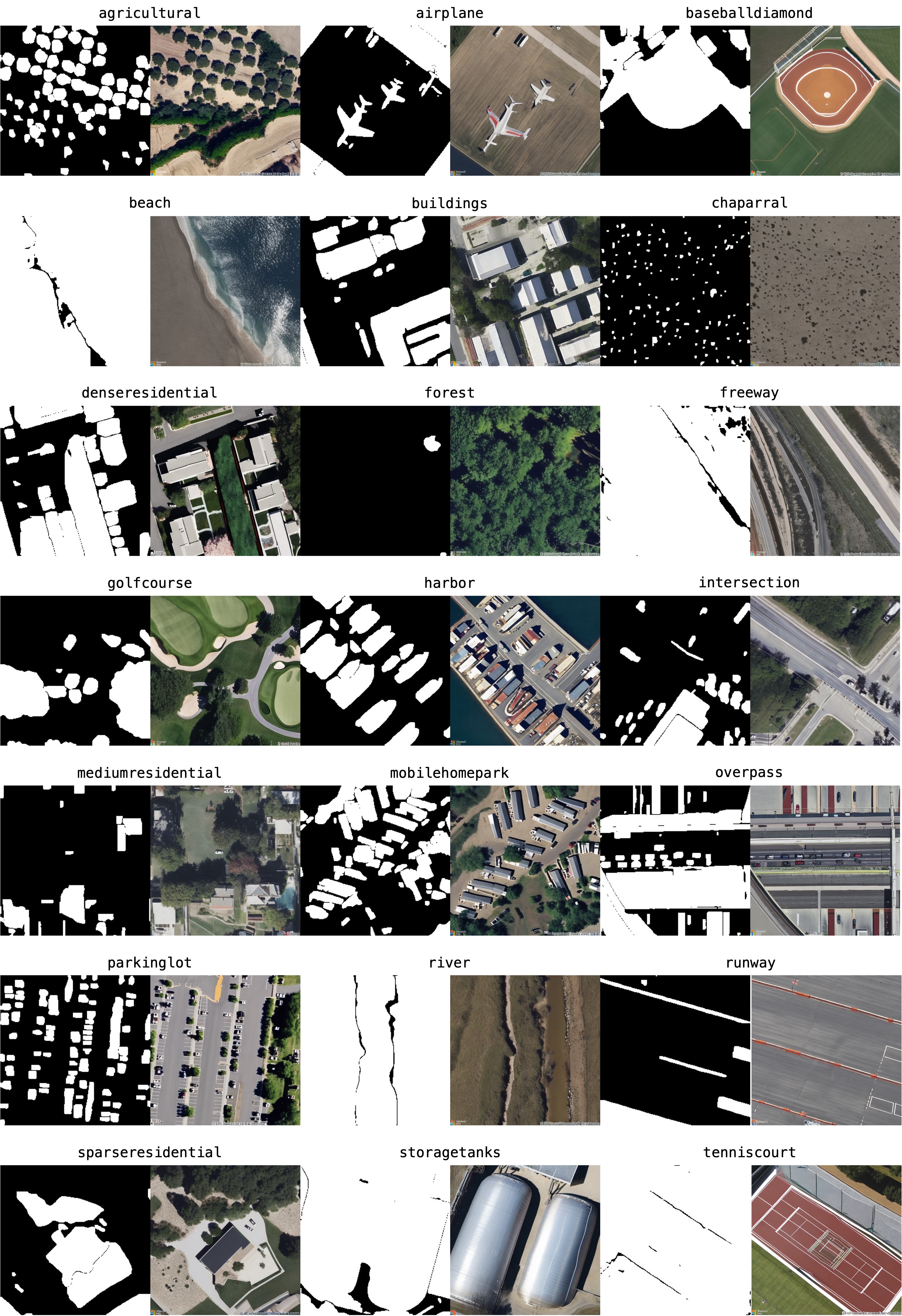}
    
    \captionof{figure}{Zero-shot synthesis using SAM mask. Examples of synthesized satellite images using SAM mask generated on UCMerced dataset. }
    \label{fig:supl_2}
\end{center}%
}]
\twocolumn[{%
\renewcommand\twocolumn[1][]{#1}%
\section{Dataset Sampling}
\label{dataset_sample}
\begin{center}
    \centering
    \captionsetup{type=figure}
    \includegraphics[width=0.5\linewidth]{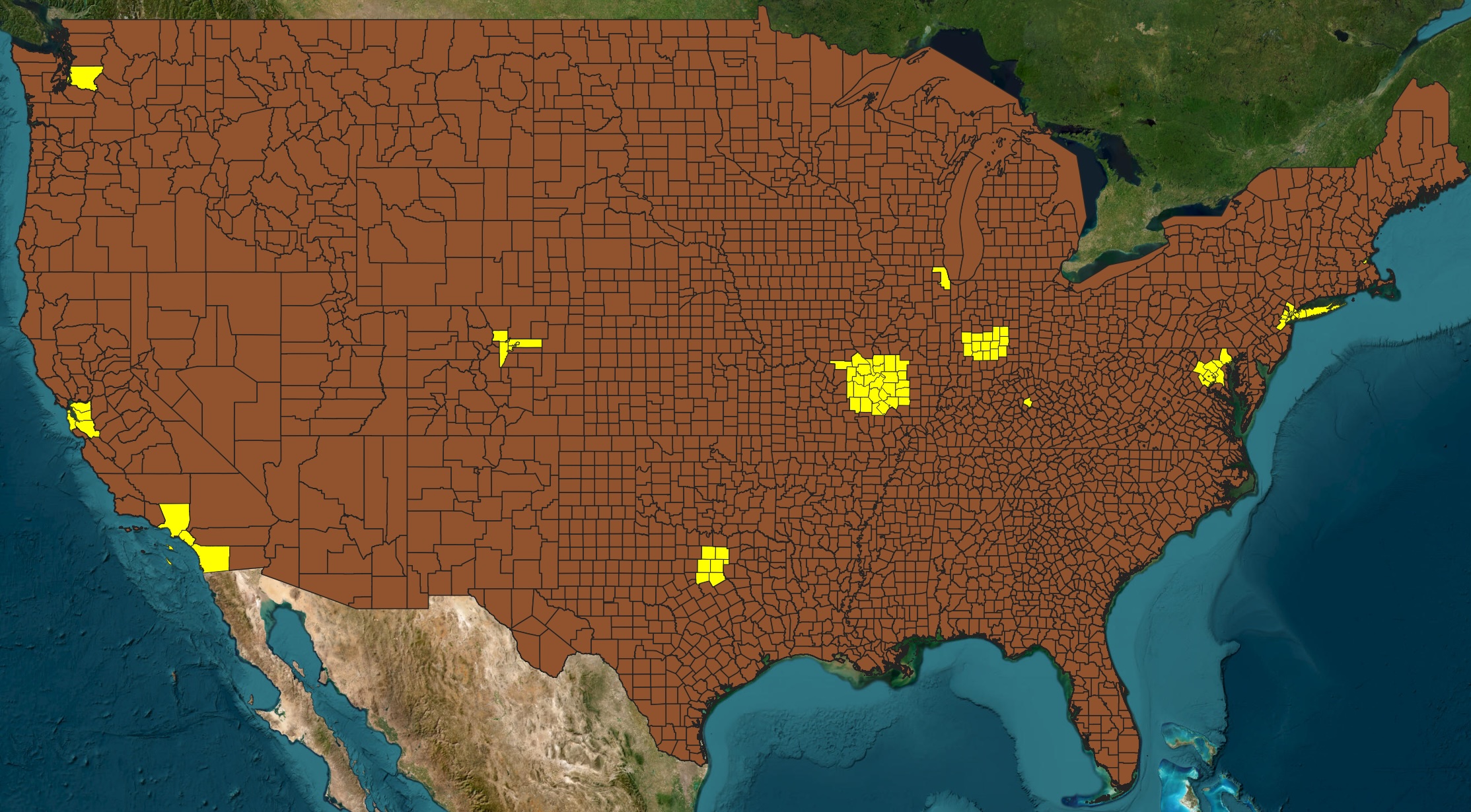}
    
    \captionof{figure}{Dataset sampling. The geographic locations across the USA used for sampling the dataset. }
    \label{fig:supl_3}
\end{center}%
\section{Limitations}
\begin{center}
    \centering
    \captionsetup{type=figure}
    \includegraphics[width=0.5\linewidth]{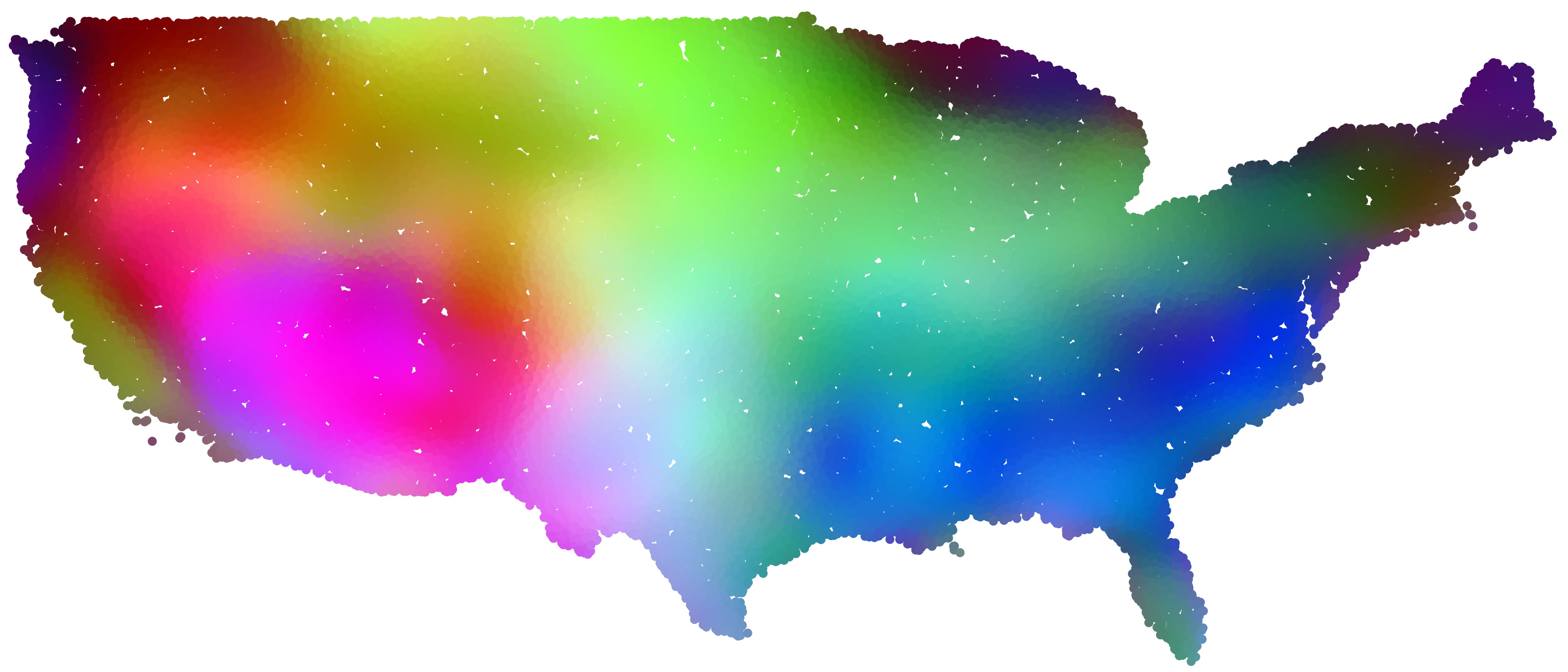}
    
    \captionof{figure}{Features Learned by SatCLIP. We show the visualization of location embeddings learned by SatCLIP. For visualization, the embeddings are projected to a 3-dimensional space using Independent Component Analysis.}
    \label{fig:supl_4}
\end{center}%
}]

%
GeoSynth is a general framework for synthesizing satellite images that combines various state-of-the-art components. As a result, its synthesis performance highly depends on the performance of each of the components. In particular, the geo-awareness of our model is restricted by the ability of SatCLIP to represent the world's geography effectively. In Figure~\ref{fig:supl_4}, we show an Independent Component Analysis (ICA) plot of SatCLIP embeddings over the USA. The figure depicts that SatCLIP cannot capture high-frequency information about the world's geography at a fine scale. Given the flexibility of our framework, SatCLIP can easily be replaced with another location encoder in the future. Currently, GeoSynth only supports synthesizing RGB-based satellite images. It is possible to extend the architecture of GeoSynth for synthesizing images coming from different modalities such as depth, radar, etc. 



%% file: main.bbl
\begin{thebibliography}{56}
\providecommand{\natexlab}[1]{#1}
\providecommand{\url}[1]{\texttt{#1}}
\expandafter\ifx\csname urlstyle\endcsname\relax
  \providecommand{\doi}[1]{doi: #1}\else
  \providecommand{\doi}{doi: \begingroup \urlstyle{rm}\Url}\fi

\bibitem[Bar-Tal et~al.(2023)Bar-Tal, Yariv, Lipman, and Dekel]{bar2023multidiffusion}
Omer Bar-Tal, Lior Yariv, Yaron Lipman, and Tali Dekel.
\newblock Multidiffusion: Fusing diffusion paths for controlled image generation.
\newblock 2023.

\bibitem[Brooks et~al.(2023)Brooks, Holynski, and Efros]{brooks2023instructpix2pix}
Tim Brooks, Aleksander Holynski, and Alexei~A Efros.
\newblock Instructpix2pix: Learning to follow image editing instructions.
\newblock In \emph{Proceedings of the IEEE/CVF Conference on Computer Vision and Pattern Recognition}, pages 18392--18402, 2023.

\bibitem[Brooks et~al.(2024)Brooks, Peebles, Homes, DePue, Guo, Jing, Schnurr, Taylor, Luhman, Luhman, Ng, Wang, and Ramesh]{videoworldsimulators2024}
Tim Brooks, Bill Peebles, Connor Homes, Will DePue, Yufei Guo, Li Jing, David Schnurr, Joe Taylor, Troy Luhman, Eric Luhman, Clarence Wing~Yin Ng, Ricky Wang, and Aditya Ramesh.
\newblock Video generation models as world simulators.
\newblock 2024.

\bibitem[Canny(1986)]{canny1986computational}
John Canny.
\newblock A computational approach to edge detection.
\newblock \emph{IEEE Transactions on pattern analysis and machine intelligence}, \penalty0 (6):\penalty0 679--698, 1986.

\bibitem[Chen et~al.(2024{\natexlab{a}})Chen, Laina, and Vedaldi]{chen2024training}
Minghao Chen, Iro Laina, and Andrea Vedaldi.
\newblock Training-free layout control with cross-attention guidance.
\newblock In \emph{Proceedings of the IEEE/CVF Winter Conference on Applications of Computer Vision}, pages 5343--5353, 2024{\natexlab{a}}.

\bibitem[Chen et~al.(2023)Chen, Yue, Fang, and Xia]{chen2023spectraldiff}
Ning Chen, Jun Yue, Leyuan Fang, and Shaobo Xia.
\newblock Spectraldiff: A generative framework for hyperspectral image classification with diffusion models.
\newblock \emph{IEEE Transactions on Geoscience and Remote Sensing}, 2023.

\bibitem[Chen et~al.(2024{\natexlab{b}})Chen, Hu, Li, Ruiz, Jia, Chang, and Cohen]{chen2024subject}
Wenhu Chen, Hexiang Hu, Yandong Li, Nataniel Ruiz, Xuhui Jia, Ming-Wei Chang, and William~W Cohen.
\newblock Subject-driven text-to-image generation via apprenticeship learning.
\newblock \emph{Advances in Neural Information Processing Systems}, 36, 2024{\natexlab{b}}.

\bibitem[Cong et~al.(2023)Cong, Khanna, Meng, Liu, Rozi, He, Burke, Lobell, and Ermon]{cong2023satmae}
Yezhen Cong, Samar Khanna, Chenlin Meng, Patrick Liu, Erik Rozi, Yutong He, Marshall Burke, David~B. Lobell, and Stefano Ermon.
\newblock Satmae: Pre-training transformers for temporal and multi-spectral satellite imagery, 2023.

\bibitem[Dhakal et~al.(2023)Dhakal, Ahmad, Khanal, Sastry, and Jacobs]{dhakal2023sat2cap}
Aayush Dhakal, Adeel Ahmad, Subash Khanal, Srikumar Sastry, and Nathan Jacobs.
\newblock Sat2cap: Mapping fine-grained textual descriptions from satellite images, 2023.

\bibitem[Epstein et~al.(2024)Epstein, Jabri, Poole, Efros, and Holynski]{epstein2024diffusion}
Dave Epstein, Allan Jabri, Ben Poole, Alexei Efros, and Aleksander Holynski.
\newblock Diffusion self-guidance for controllable image generation.
\newblock \emph{Advances in Neural Information Processing Systems}, 36, 2024.

\bibitem[Esser et~al.(2023)Esser, Chiu, Atighehchian, Granskog, and Germanidis]{esser2023structure}
Patrick Esser, Johnathan Chiu, Parmida Atighehchian, Jonathan Granskog, and Anastasis Germanidis.
\newblock Structure and content-guided video synthesis with diffusion models.
\newblock In \emph{Proceedings of the IEEE/CVF International Conference on Computer Vision}, pages 7346--7356, 2023.

\bibitem[Gal et~al.(2022)Gal, Alaluf, Atzmon, Patashnik, Bermano, Chechik, and Cohen-Or]{gal2022image}
Rinon Gal, Yuval Alaluf, Yuval Atzmon, Or Patashnik, Amit~H Bermano, Gal Chechik, and Daniel Cohen-Or.
\newblock An image is worth one word: Personalizing text-to-image generation using textual inversion.
\newblock \emph{arXiv preprint arXiv:2208.01618}, 2022.

\bibitem[Guo et~al.(2023)Guo, Yang, Rao, Wang, Qiao, Lin, and Dai]{guo2023animatediff}
Yuwei Guo, Ceyuan Yang, Anyi Rao, Yaohui Wang, Yu Qiao, Dahua Lin, and Bo Dai.
\newblock Animatediff: Animate your personalized text-to-image diffusion models without specific tuning.
\newblock \emph{arXiv preprint arXiv:2307.04725}, 2023.

\bibitem[Gupta et~al.(2023)Gupta, Yu, Sohn, Gu, Hahn, Fei-Fei, Essa, Jiang, and Lezama]{gupta2023photorealistic}
Agrim Gupta, Lijun Yu, Kihyuk Sohn, Xiuye Gu, Meera Hahn, Li Fei-Fei, Irfan Essa, Lu Jiang, and Jos{\'e} Lezama.
\newblock Photorealistic video generation with diffusion models.
\newblock \emph{arXiv preprint arXiv:2312.06662}, 2023.

\bibitem[He et~al.(2021)He, Wang, Lai, Zhang, Meng, Burke, Lobell, and Ermon]{he2021spatial}
Yutong He, Dingjie Wang, Nicholas Lai, William Zhang, Chenlin Meng, Marshall Burke, David Lobell, and Stefano Ermon.
\newblock Spatial-temporal super-resolution of satellite imagery via conditional pixel synthesis.
\newblock \emph{Advances in Neural Information Processing Systems}, 34:\penalty0 27903--27915, 2021.

\bibitem[Hertz et~al.(2022)Hertz, Mokady, Tenenbaum, Aberman, Pritch, and Cohen-Or]{hertz2022prompt}
Amir Hertz, Ron Mokady, Jay Tenenbaum, Kfir Aberman, Yael Pritch, and Daniel Cohen-Or.
\newblock Prompt-to-prompt image editing with cross attention control.
\newblock \emph{arXiv preprint arXiv:2208.01626}, 2022.

\bibitem[Heusel et~al.(2017)Heusel, Ramsauer, Unterthiner, Nessler, and Hochreiter]{heusel2017gans}
Martin Heusel, Hubert Ramsauer, Thomas Unterthiner, Bernhard Nessler, and Sepp Hochreiter.
\newblock Gans trained by a two time-scale update rule converge to a local nash equilibrium.
\newblock \emph{Advances in neural information processing systems}, 30, 2017.

\bibitem[Ho et~al.(2020)Ho, Jain, and Abbeel]{ho2020denoising}
Jonathan Ho, Ajay Jain, and Pieter Abbeel.
\newblock Denoising diffusion probabilistic models.
\newblock \emph{Advances in neural information processing systems}, 33:\penalty0 6840--6851, 2020.

\bibitem[Huang et~al.(2024)Huang, Zhang, Tang, Ma, Huang, Dong, and Xu]{huang2024diffstyler}
Nisha Huang, Yuxin Zhang, Fan Tang, Chongyang Ma, Haibin Huang, Weiming Dong, and Changsheng Xu.
\newblock Diffstyler: Controllable dual diffusion for text-driven image stylization.
\newblock \emph{IEEE Transactions on Neural Networks and Learning Systems}, 2024.

\bibitem[Jakubik et~al.(2023)Jakubik, Roy, Phillips, Fraccaro, Godwin, Zadrozny, Szwarcman, Gomes, Nyirjesy, Edwards, Kimura, Simumba, Chu, Mukkavilli, Lambhate, Das, Bangalore, Oliveira, Muszynski, Ankur, Ramasubramanian, Gurung, Khallaghi, Hanxi, Li, Cecil, Ahmadi, Kordi, Alemohammad, Maskey, Ganti, Weldemariam, and Ramachandran]{jakubik2023foundation}
Johannes Jakubik, Sujit Roy, C.~E. Phillips, Paolo Fraccaro, Denys Godwin, Bianca Zadrozny, Daniela Szwarcman, Carlos Gomes, Gabby Nyirjesy, Blair Edwards, Daiki Kimura, Naomi Simumba, Linsong Chu, S.~Karthik Mukkavilli, Devyani Lambhate, Kamal Das, Ranjini Bangalore, Dario Oliveira, Michal Muszynski, Kumar Ankur, Muthukumaran Ramasubramanian, Iksha Gurung, Sam Khallaghi, Hanxi, Li, Michael Cecil, Maryam Ahmadi, Fatemeh Kordi, Hamed Alemohammad, Manil Maskey, Raghu Ganti, Kommy Weldemariam, and Rahul Ramachandran.
\newblock Foundation models for generalist geospatial artificial intelligence, 2023.

\bibitem[Kawar et~al.(2023)Kawar, Zada, Lang, Tov, Chang, Dekel, Mosseri, and Irani]{kawar2023imagic}
Bahjat Kawar, Shiran Zada, Oran Lang, Omer Tov, Huiwen Chang, Tali Dekel, Inbar Mosseri, and Michal Irani.
\newblock Imagic: Text-based real image editing with diffusion models.
\newblock In \emph{Proceedings of the IEEE/CVF Conference on Computer Vision and Pattern Recognition}, pages 6007--6017, 2023.

\bibitem[Khanal et~al.(2023)Khanal, Sastry, Dhakal, and Jacobs]{khanal2023learning}
Subash Khanal, Srikumar Sastry, Aayush Dhakal, and Nathan Jacobs.
\newblock Learning tri-modal embeddings for zero-shot soundscape mapping, 2023.

\bibitem[Khanna et~al.(2023)Khanna, Liu, Zhou, Meng, Rombach, Burke, Lobell, and Ermon]{khanna2023diffusionsat}
Samar Khanna, Patrick Liu, Linqi Zhou, Chenlin Meng, Robin Rombach, Marshall Burke, David Lobell, and Stefano Ermon.
\newblock Diffusionsat: A generative foundation model for satellite imagery.
\newblock \emph{arXiv preprint arXiv:2312.03606}, 2023.

\bibitem[Kirillov et~al.(2023)Kirillov, Mintun, Ravi, Mao, Rolland, Gustafson, Xiao, Whitehead, Berg, Lo, et~al.]{kirillov2023segment}
Alexander Kirillov, Eric Mintun, Nikhila Ravi, Hanzi Mao, Chloe Rolland, Laura Gustafson, Tete Xiao, Spencer Whitehead, Alexander~C Berg, Wan-Yen Lo, et~al.
\newblock Segment anything.
\newblock \emph{arXiv preprint arXiv:2304.02643}, 2023.

\bibitem[Klemmer et~al.(2023)Klemmer, Rolf, Robinson, Mackey, and Ru{\ss}wurm]{klemmer2023satclip}
Konstantin Klemmer, Esther Rolf, Caleb Robinson, Lester Mackey, and Marc Ru{\ss}wurm.
\newblock Satclip: Global, general-purpose location embeddings with satellite imagery.
\newblock \emph{arXiv preprint arXiv:2311.17179}, 2023.

\bibitem[Kumari et~al.(2023)Kumari, Zhang, Zhang, Shechtman, and Zhu]{kumari2023multi}
Nupur Kumari, Bingliang Zhang, Richard Zhang, Eli Shechtman, and Jun-Yan Zhu.
\newblock Multi-concept customization of text-to-image diffusion.
\newblock In \emph{Proceedings of the IEEE/CVF Conference on Computer Vision and Pattern Recognition}, pages 1931--1941, 2023.

\bibitem[Li et~al.(2024)Li, Li, and Hoi]{li2024blip}
Dongxu Li, Junnan Li, and Steven Hoi.
\newblock Blip-diffusion: Pre-trained subject representation for controllable text-to-image generation and editing.
\newblock \emph{Advances in Neural Information Processing Systems}, 36, 2024.

\bibitem[Liu et~al.(2024)Liu, Li, Wu, and Lee]{liu2024visual}
Haotian Liu, Chunyuan Li, Qingyang Wu, and Yong~Jae Lee.
\newblock Visual instruction tuning.
\newblock \emph{Advances in neural information processing systems}, 36, 2024.

\bibitem[Lu et~al.(2023)Lu, Tunanyan, Wang, Navasardyan, Wang, and Shi]{lu2023specialist}
Haoming Lu, Hazarapet Tunanyan, Kai Wang, Shant Navasardyan, Zhangyang Wang, and Humphrey Shi.
\newblock Specialist diffusion: Plug-and-play sample-efficient fine-tuning of text-to-image diffusion models to learn any unseen style.
\newblock In \emph{Proceedings of the IEEE/CVF Conference on Computer Vision and Pattern Recognition}, pages 14267--14276, 2023.

\bibitem[Nitzan et~al.(2023)Nitzan, Gharbi, Zhang, Park, Zhu, Cohen-Or, and Shechtman]{nitzan2023domain}
Yotam Nitzan, Micha{\"e}l Gharbi, Richard Zhang, Taesung Park, Jun-Yan Zhu, Daniel Cohen-Or, and Eli Shechtman.
\newblock Domain expansion of image generators.
\newblock In \emph{Proceedings of the IEEE/CVF Conference on Computer Vision and Pattern Recognition}, pages 15933--15942, 2023.

\bibitem[Radford et~al.(2021)Radford, Kim, Hallacy, Ramesh, Goh, Agarwal, Sastry, Askell, Mishkin, Clark, et~al.]{radford2021learning}
Alec Radford, Jong~Wook Kim, Chris Hallacy, Aditya Ramesh, Gabriel Goh, Sandhini Agarwal, Girish Sastry, Amanda Askell, Pamela Mishkin, Jack Clark, et~al.
\newblock Learning transferable visual models from natural language supervision.
\newblock In \emph{International conference on machine learning}, pages 8748--8763. PMLR, 2021.

\bibitem[Ramesh et~al.(2022)Ramesh, Dhariwal, Nichol, Chu, and Chen]{ramesh2022hierarchical}
Aditya Ramesh, Prafulla Dhariwal, Alex Nichol, Casey Chu, and Mark Chen.
\newblock Hierarchical text-conditional image generation with clip latents.
\newblock \emph{arXiv preprint arXiv:2204.06125}, 1\penalty0 (2):\penalty0 3, 2022.

\bibitem[Reed et~al.(2023)Reed, Gupta, Li, Brockman, Funk, Clipp, Keutzer, Candido, Uyttendaele, and Darrell]{reed2023scalemae}
Colorado~J. Reed, Ritwik Gupta, Shufan Li, Sarah Brockman, Christopher Funk, Brian Clipp, Kurt Keutzer, Salvatore Candido, Matt Uyttendaele, and Trevor Darrell.
\newblock Scale-mae: A scale-aware masked autoencoder for multiscale geospatial representation learning, 2023.

\bibitem[Rolf et~al.(2024)Rolf, Klemmer, Robinson, and Kerner]{Rolf2024MissionC}
Esther Rolf, Konstantin Klemmer, Caleb Robinson, and Hannah Kerner.
\newblock Mission critical - satellite data is a distinct modality in machine learning.
\newblock \emph{ArXiv}, abs/2402.01444, 2024.

\bibitem[Rombach et~al.(2022)Rombach, Blattmann, Lorenz, Esser, and Ommer]{rombach2022high}
Robin Rombach, Andreas Blattmann, Dominik Lorenz, Patrick Esser, and Bj{\"o}rn Ommer.
\newblock High-resolution image synthesis with latent diffusion models.
\newblock In \emph{Proceedings of the IEEE/CVF conference on computer vision and pattern recognition}, pages 10684--10695, 2022.

\bibitem[Ruiz et~al.(2023{\natexlab{a}})Ruiz, Li, Jampani, Pritch, Rubinstein, and Aberman]{ruiz2023dreambooth}
Nataniel Ruiz, Yuanzhen Li, Varun Jampani, Yael Pritch, Michael Rubinstein, and Kfir Aberman.
\newblock Dreambooth: Fine tuning text-to-image diffusion models for subject-driven generation.
\newblock In \emph{Proceedings of the IEEE/CVF Conference on Computer Vision and Pattern Recognition}, pages 22500--22510, 2023{\natexlab{a}}.

\bibitem[Ruiz et~al.(2023{\natexlab{b}})Ruiz, Li, Jampani, Wei, Hou, Pritch, Wadhwa, Rubinstein, and Aberman]{ruiz2023hyperdreambooth}
Nataniel Ruiz, Yuanzhen Li, Varun Jampani, Wei Wei, Tingbo Hou, Yael Pritch, Neal Wadhwa, Michael Rubinstein, and Kfir Aberman.
\newblock Hyperdreambooth: Hypernetworks for fast personalization of text-to-image models.
\newblock \emph{arXiv preprint arXiv:2307.06949}, 2023{\natexlab{b}}.

\bibitem[Saharia et~al.(2022)Saharia, Chan, Saxena, Li, Whang, Denton, Ghasemipour, Gontijo~Lopes, Karagol~Ayan, Salimans, et~al.]{saharia2022photorealistic}
Chitwan Saharia, William Chan, Saurabh Saxena, Lala Li, Jay Whang, Emily~L Denton, Kamyar Ghasemipour, Raphael Gontijo~Lopes, Burcu Karagol~Ayan, Tim Salimans, et~al.
\newblock Photorealistic text-to-image diffusion models with deep language understanding.
\newblock \emph{Advances in Neural Information Processing Systems}, 35:\penalty0 36479--36494, 2022.

\bibitem[Silva et~al.(2024)Silva, Magalhães, Tuia, and Martins]{silva2024large}
João~Daniel Silva, João Magalhães, Devis Tuia, and Bruno Martins.
\newblock Large language models for captioning and retrieving remote sensing images, 2024.

\bibitem[Sohl-Dickstein et~al.(2015)Sohl-Dickstein, Weiss, Maheswaranathan, and Ganguli]{sohl2015deep}
Jascha Sohl-Dickstein, Eric Weiss, Niru Maheswaranathan, and Surya Ganguli.
\newblock Deep unsupervised learning using nonequilibrium thermodynamics.
\newblock In \emph{International conference on machine learning}, pages 2256--2265. PMLR, 2015.

\bibitem[Sohn et~al.(2024)Sohn, Jiang, Barber, Lee, Ruiz, Krishnan, Chang, Li, Essa, Rubinstein, et~al.]{sohn2024styledrop}
Kihyuk Sohn, Lu Jiang, Jarred Barber, Kimin Lee, Nataniel Ruiz, Dilip Krishnan, Huiwen Chang, Yuanzhen Li, Irfan Essa, Michael Rubinstein, et~al.
\newblock Styledrop: Text-to-image synthesis of any style.
\newblock \emph{Advances in Neural Information Processing Systems}, 36, 2024.

\bibitem[Tseng et~al.(2024)Tseng, Cartuyvels, Zvonkov, Purohit, Rolnick, and Kerner]{tseng2024lightweight}
Gabriel Tseng, Ruben Cartuyvels, Ivan Zvonkov, Mirali Purohit, David Rolnick, and Hannah Kerner.
\newblock Lightweight, pre-trained transformers for remote sensing timeseries, 2024.

\bibitem[Tumanyan et~al.(2023)Tumanyan, Geyer, Bagon, and Dekel]{tumanyan2023plug}
Narek Tumanyan, Michal Geyer, Shai Bagon, and Tali Dekel.
\newblock Plug-and-play diffusion features for text-driven image-to-image translation.
\newblock In \emph{Proceedings of the IEEE/CVF Conference on Computer Vision and Pattern Recognition}, pages 1921--1930, 2023.

\bibitem[Van Den~Oord et~al.(2017)Van Den~Oord, Vinyals, et~al.]{van2017neural}
Aaron Van Den~Oord, Oriol Vinyals, et~al.
\newblock Neural discrete representation learning.
\newblock \emph{Advances in neural information processing systems}, 30, 2017.

\bibitem[Vivanco et~al.(2023)Vivanco, Nayak, and Shah]{cepeda2023geoclip}
Vicente Vivanco, Gaurav~Kumar Nayak, and Mubarak Shah.
\newblock Geoclip: Clip-inspired alignment between locations and images for effective worldwide geo-localization.
\newblock 2023.

\bibitem[Wang et~al.(2004)Wang, Bovik, Sheikh, and Simoncelli]{wang2004image}
Zhou Wang, Alan~C Bovik, Hamid~R Sheikh, and Eero~P Simoncelli.
\newblock Image quality assessment: from error visibility to structural similarity.
\newblock \emph{IEEE transactions on image processing}, 13\penalty0 (4):\penalty0 600--612, 2004.

\bibitem[Xiao et~al.(2023)Xiao, Yuan, Jiang, He, Jin, and Zhang]{xiao2023ediffsr}
Yi Xiao, Qiangqiang Yuan, Kui Jiang, Jiang He, Xianyu Jin, and Liangpei Zhang.
\newblock Ediffsr: An efficient diffusion probabilistic model for remote sensing image super-resolution.
\newblock \emph{IEEE Transactions on Geoscience and Remote Sensing}, 2023.

\bibitem[Xie et~al.(2023)Xie, Li, Huang, Liu, Zhang, Zheng, and Shou]{xie2023boxdiff}
Jinheng Xie, Yuexiang Li, Yawen Huang, Haozhe Liu, Wentian Zhang, Yefeng Zheng, and Mike~Zheng Shou.
\newblock Boxdiff: Text-to-image synthesis with training-free box-constrained diffusion.
\newblock In \emph{Proceedings of the IEEE/CVF International Conference on Computer Vision}, pages 7452--7461, 2023.

\bibitem[Xu et~al.(2024)Xu, Motamed, Vaddamanu, Wu, Haene, Bazin, and De~la Torre]{xu2024personalized}
Jianjin Xu, Saman Motamed, Praneetha Vaddamanu, Chen~Henry Wu, Christian Haene, Jean-Charles Bazin, and Fernando De~la Torre.
\newblock Personalized face inpainting with diffusion models by parallel visual attention.
\newblock In \emph{Proceedings of the IEEE/CVF Winter Conference on Applications of Computer Vision}, pages 5432--5442, 2024.

\bibitem[Xu et~al.(2023)Xu, Yu, Ghamisi, Kopp, and Hochreiter]{xu2023txt2img}
Yonghao Xu, Weikang Yu, Pedram Ghamisi, Michael Kopp, and Sepp Hochreiter.
\newblock Txt2img-mhn: Remote sensing image generation from text using modern hopfield networks.
\newblock \emph{IEEE Transactions on Image Processing}, 2023.

\bibitem[Xue et~al.(2023)Xue, Huang, Sun, Song, and Zhang]{xue2023freestyle}
Han Xue, Zhiwu Huang, Qianru Sun, Li Song, and Wenjun Zhang.
\newblock Freestyle layout-to-image synthesis.
\newblock In \emph{Proceedings of the IEEE/CVF Conference on Computer Vision and Pattern Recognition}, pages 14256--14266, 2023.

\bibitem[Yang and Newsam(2010)]{yang2010bag}
Yi Yang and Shawn Newsam.
\newblock Bag-of-visual-words and spatial extensions for land-use classification.
\newblock In \emph{Proceedings of the 18th SIGSPATIAL international conference on advances in geographic information systems}, pages 270--279, 2010.

\bibitem[Zavras et~al.(2024)Zavras, Michail, Demir, and Papoutsis]{zavras2024mind}
Angelos Zavras, Dimitrios Michail, Beg{\"u}m Demir, and Ioannis Papoutsis.
\newblock Mind the modality gap: Towards a remote sensing vision-language model via cross-modal alignment.
\newblock \emph{arXiv preprint arXiv:2402.09816}, 2024.

\bibitem[Zhang et~al.(2023)Zhang, Rao, and Agrawala]{zhang2023adding}
Lvmin Zhang, Anyi Rao, and Maneesh Agrawala.
\newblock Adding conditional control to text-to-image diffusion models.
\newblock In \emph{Proceedings of the IEEE/CVF International Conference on Computer Vision}, pages 3836--3847, 2023.

\bibitem[Zhang et~al.(2024)Zhang, Cai, Zhang, Zhuang, and Mao]{zhang2024earthgpt}
Wei Zhang, Miaoxin Cai, Tong Zhang, Yin Zhuang, and Xuerui Mao.
\newblock Earthgpt: A universal multi-modal large language model for multi-sensor image comprehension in remote sensing domain, 2024.

\bibitem[Zhao et~al.(2024)Zhao, Chen, Chen, Bao, Hao, Yuan, and Wong]{zhao2024uni}
Shihao Zhao, Dongdong Chen, Yen-Chun Chen, Jianmin Bao, Shaozhe Hao, Lu Yuan, and Kwan-Yee~K Wong.
\newblock Uni-controlnet: All-in-one control to text-to-image diffusion models.
\newblock \emph{Advances in Neural Information Processing Systems}, 36, 2024.

\end{thebibliography}
